\documentclass[10pt,twocolumn,letterpaper]{article}

\usepackage{cvpr}              

\usepackage{colortbl}
\newcommand{\ours}{FRAG\xspace}
\usepackage[normalem]{ulem}
\useunder{\uline}{\ul}{}
\usepackage{pifont}
\newcommand{\cmark}{\ding{51}}
\newcommand{\xmark}{\ding{55}}
\newcommand{\tabsize}{\small}

\definecolor{cvprblue}{rgb}{0.21,0.49,0.74}
\usepackage[pagebackref,breaklinks,colorlinks,allcolors=cvprblue]{hyperref}

\def\paperID{****} 
\def\confName{CVPR}
\def\confYear{2025}

\title{FRAG: Frame Selection Augmented Generation for Long Video and Long Document Understanding}

\author{
De-An Huang,
Subhashree Radhakrishnan,
Zhiding Yu,
Jan Kautz \\
NVIDIA \\
{\tt\small \{deahuang,subhashreer,zhidingy,jkautz\}@nvidia.com}
}

\begin{document}
\maketitle
\begin{abstract}
There has been impressive progress in Large Multimodal Models (LMMs). Recent works extend these models to long inputs, including multi-page documents and long videos. However, the model size and performance of these long context models are still limited due to the computational cost in both training and inference. In this work, we explore an orthogonal direction and process long inputs without long context LMMs. We propose Frame Selection Augmented Generation (\ours), where the model first selects relevant frames within the input, and then only generates the final outputs based on the selected frames. The core of the selection process is done by scoring each frame independently, which does not require long context processing. The frames with the highest scores are then selected by a simple Top-K selection. We show that this frustratingly simple framework is applicable to both long videos and multi-page documents using existing LMMs without any fine-tuning. We consider two models, LLaVA-OneVision and InternVL2, in our experiments and show that \ours  consistently improves the performance and achieves state-of-the-art performances for both long video and long document understanding. For videos, \ours substantially improves InternVL2-76B by 5.8\% on MLVU and 3.7\% on Video-MME. For documents, \ours achieves over 20\% improvements on MP-DocVQA compared with recent LMMs specialized in long document understanding. Code is available at: \url{https://github.com/NVlabs/FRAG}
\end{abstract}    
\section{Introduction}
\label{sec:intro}

Large Multimodal Models (LMMs) have demonstrated impressive multimodal understanding capabilities such as captioning and visual question answering on images and videos~\cite{chen2024far,li2024llava,liu2023llava,Qwen-VL}. Recent works further extend LMMs to long input data, such as multi-page documents and long videos~\cite{hu2024mplug,xue2024longvila,zhang2024longva}.

Training these long context LMMs, however, poses challenges to both data and computation. First, high-quality long context data is even more costly to obtain than typical multimodal data. Recent works thus aim to train long context LMMs without or with only limited long context multimodal data~\cite{zhang2024longva}.  
Second, long context LMMs have much higher computational costs. 
Even with sequence parallelism methods to avoid memory limitation of a single device, the computation requirement is still much higher than typical training of LMMs~\cite{xue2024longvila}. Therefore, existing long context LMMs are limited in model sizes -- often in the 7B range. This is in contrast to typical LMMs, which commonly scale to over 70B~\cite{chen2024far,li2024llava}. Both of these challenges limit the capabilities of existing long context LMMs .

\begin{figure}[t]
  \centering
  \includegraphics[width=1.0\linewidth]{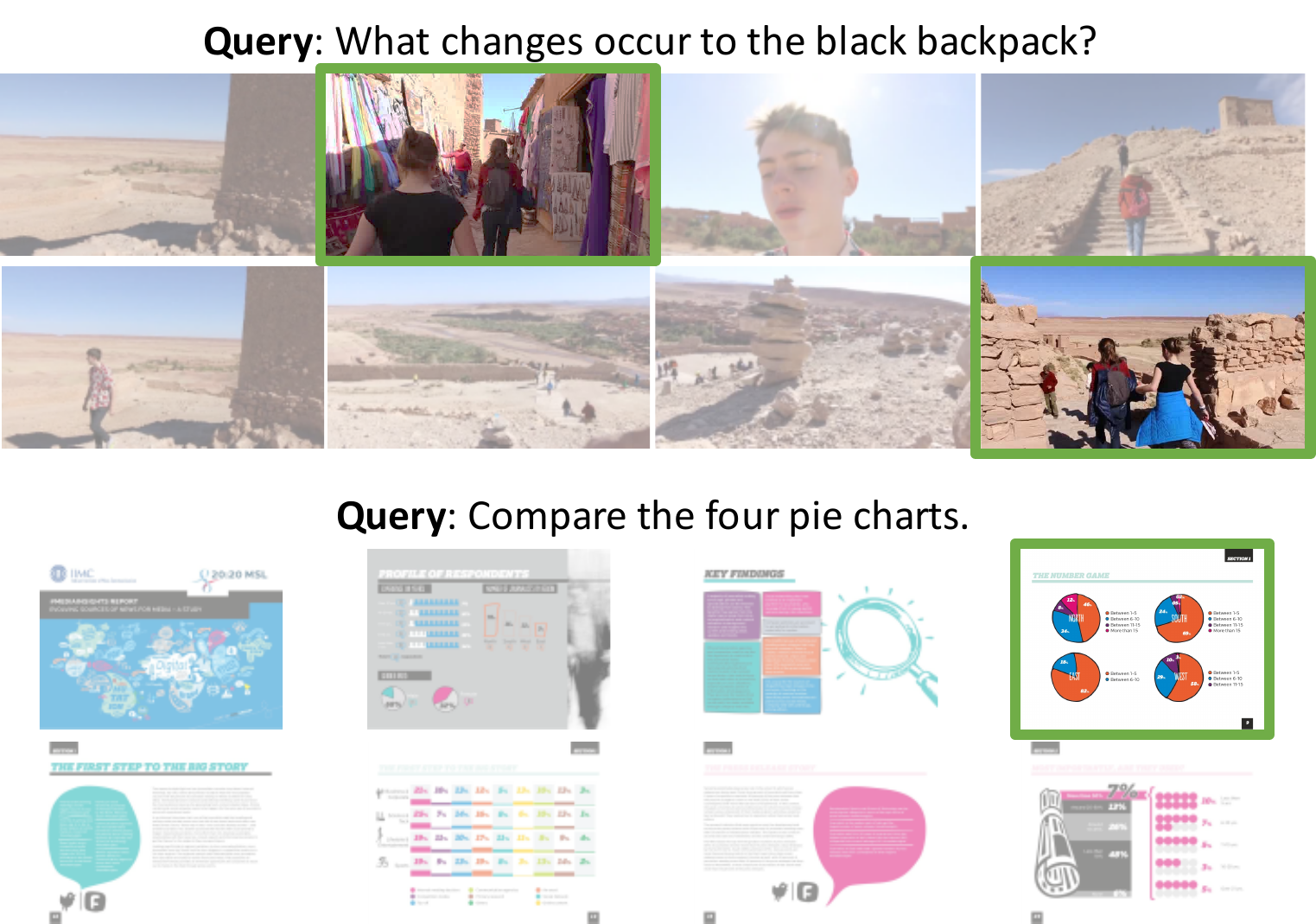}
  \caption{Questions about long inputs can often be answered without long context global processing. For the query about the black backpack in the video, one would focus on the frames with black backpacks. Similarly for the slides, one would just focus on the pages with pie charts if the question is about pie charts. }
  \label{fig:fig1}
\end{figure}

In this work, we explore an orthogonal direction and ask the following question: \emph{``Do we need long context LMMs for long input data?''} For example in \Cref{fig:fig1}, 
to answer a question about the black backpack in the long video, the first step is to fast-forward through the video and find the relevant moments to answer the question (\ie frames with the black backpack). Similarly for long documents, to answer a question about pie charts in the slides in \Cref{fig:fig1}, one would again first skim through the pages, and only focus on the  pages with pie charts.  
In these examples, the whole long input is not processed at once. Instead, a decision is made for each frame/page independently.
To determine whether a frame/page is relevant to the question, one can simply check whether relevant information is presented in the frame/page, which does not require global long context processing.

\textbf{Our Approach.} Based on this observation, we propose \textbf{Fr}ame Selection \textbf{A}ugmented \textbf{G}eneration (\ours) for long video and long document understanding. Given a long input and a query, \ours first scores each frame/page independently for its relevance to the query. The frames/pages with the highest scores are then selected and feed into a LMM to generate the final outputs. Since the frames/pages are scored independently, \ours does not require long-context LMMs to process the whole input.

The important technical question, however, is how the frames/pages are scored and thus selected in \ours. Our key observation is that existing LMMs can perform \emph{zero-shot} scoring without any further tuning. For each frame, we ask LMMs ``Does this image contain sufficient information to answer the query?'' and the probability that the LMMs answer ``yes'' to this prompt is highly effective for selecting the relevant frames. This observation is applicable to a surprisingly wide range of LMMs.

This leads to a frustratingly simple yet effective inference procedure for \ours. Given a LMM, \ours answers a query about long inputs by: (1) scoring each frame in the input independently by asking the LMM whether the frame contains sufficient information to answer the query. (2) Select the Top-K frames by the probability that the LMM answers yes, and feed the selected frames back to the LMM to generate the final answer.

\ours thus addresses both of the challenges of existing methods for long context LMMs. First, \ours is zero-shot and thus does not need long context training data. Second, \ours is applicable to any model size, and is not limited to the smaller model sizes because of computational limitations. Computationally, \ours is also more affordable and flexible than processing the same amount of frames jointly by a long context LMM because the quadratic self-attention never processes all the frames at once. \ours's scoring process can be easily parallelized across different devices as each frame is scored independently.

We show that \ours consistently improves the performance for long video and long document understanding for two LMM families and five model sizes~\cite{chen2024far,li2024llava}. \ours unifies long video and long document understanding, and achieves state-of-the-art results on five long video benchmarks and three multi-page document benchmarks. For videos, \ours substantially improves InternVL2-76B by 5.8\% on MLVU~\cite{MLVU} and 3.7\% on Video-MME~\cite{fu2024video}. On LongVideoBench~\cite{wu2024longvideobench}, \ours even outperforms GPT-4o. For documents, \ours outperforms specialist models that require training and OCR modules, doubling its F1 score on SlideVQA~\cite{SlideVQA2023}. \ours also outperforms recent LMMs specialized in long document understanding by over 20\% on MP-DocVQA~\cite{tito2023hierarchical}.

\section{Related Work}
\label{sec:related}

\noindent\textbf{Long Video Understanding.} 
With the rise of long context LMMs, many long video evaluation datasets have recently emerged~\cite{wang2024lvbench,wu2024longvideobench,MLVU}. 
One approach to address long video understanding is to compress the visual information through memory or resampler~\cite{balavzevic2024memory,he2024ma,korbar2023text,qian2024streaming,shen2024longvu,weng2024longvlm}. The memory or resampler is often jointly trained with the LMMs. In contrast, one benefit of \ours is that the answering LMM can leverage the selected frames without further tuning. 
Another approach for long video understanding, is to focus on extending the context length of LMMs. While we have seen promising results on processing hundreds of frames, the model size is often limited due to the computational burden introduced by the long context~\cite{xue2024longvila,zhang2024longva}. This in turn limits the performance of these models. \ours processes long videos and long documents without long context modeling and thus does not have the limitation.

\vspace{1mm}
\noindent\textbf{Multi-Page Document Understanding.} While there have been specialist models designed for multi-page documents~\cite{blau2024gram,SlideVQA2023,tito2023hierarchical}, LMMs for document understanding are mostly limited to single page documents~\cite{chen2024far,dong2024internlm,hu2024mplug15}. In contrast to specialist models, which use OCR, object detection, and fine-grained supervision, LMMs for documents  are often OCR-free and directly process the whole page as an image. However, this makes multi-page document understanding challenging as the number of tokens significantly increases for high-resolution multi-page inputs. Concurrent works on LMMs for multi-page documents thus focus on adaptive high-resolution modules that dynamically allocate visual tokens based on document layouts~\cite{hu2024mplug,jia2024leopard}. \ours's frame selection significantly reduces the number of visual tokens, and can easily apply high resolution visual processing to areas that are relevant to the question.

\vspace{1mm}
\noindent\textbf{Frame Selection for Large Multimodal Models.} Frame selection has a long history in video understanding~\cite{korbar2019scsampler}. Recent LMM works that use frame selection often first convert video frames to text and depend on proprietary language models~\cite{park2024too,wang2024videoagent,wang2024videotree}. The most related to our work is SeViLA, which trains a frame selector using LMMs~\cite{yu2024self}. Concurrent works also explore visual content selection for long videos~\cite{yu2024frame}, PDF documents~\cite{xie2024wukong}, and multi-image question answering~\cite{wu2024visual}. These works require training for selection and are application specific. In contrast, \ours is zero-shot and can be applied to any of the applications.

\begin{figure*}[t]
  \centering
  \includegraphics[width=1.0\linewidth]{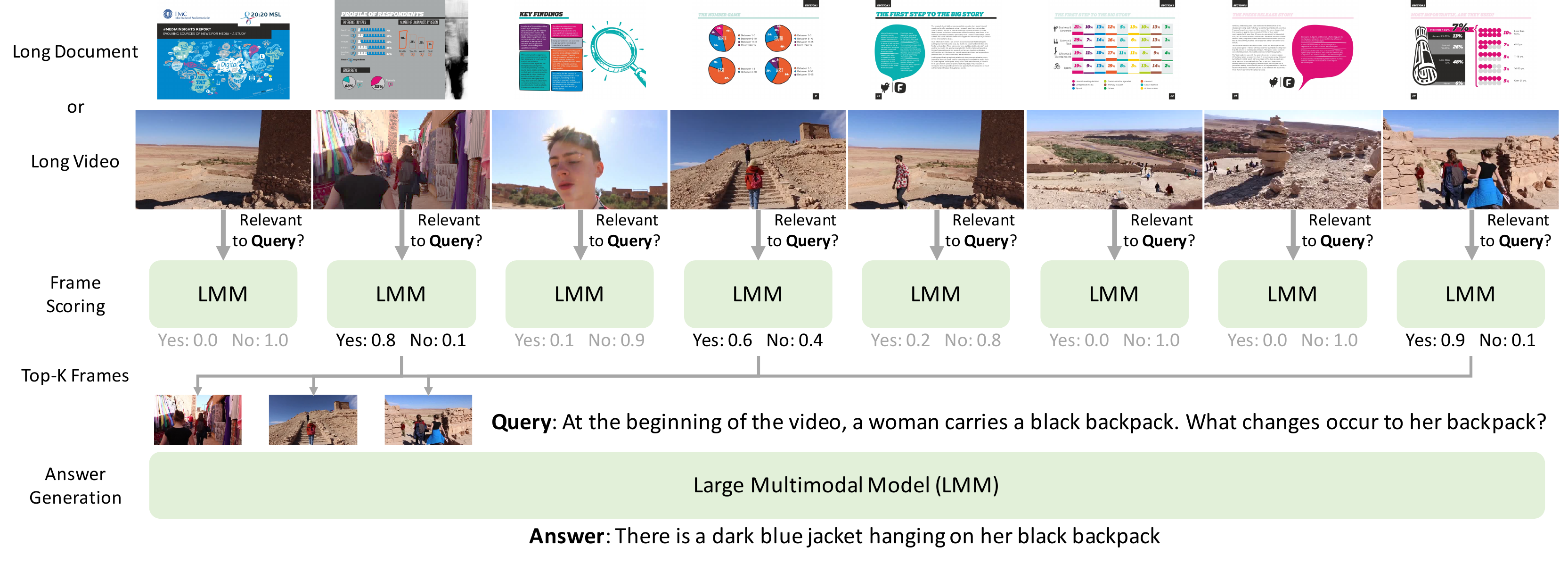}
  \caption{Overview of \ours. \ours first uses a scoring LMM to score each sampled frame in a video or document. The Top-K scoring frames are then selected to use as input to the answering LMM for answer generation. The scoring LMM and the answering LMM can be the same, but are not required to be the same. We find that existing LMMs can serve both purposes without any tuning.}
  \label{fig:system}
\end{figure*}
\section{Frame Selection Augmented Generation}
\label{sec:method}

\ours process long videos and long documents by first selecting the relevant frames/pages. The selection process is done by (1) scoring each frame independently, and (2) selecting the top scoring frames to generate the final outputs. Since the scoring is done independently, \ours does not require long context to process long videos and documents. \Cref{fig:system} shows an overview of \ours.

\subsection{Independent Frame Scoring}

Given a long video or document, the first step of \ours is to score selection proposals. This consists of two steps: proposal generation and proposal scoring. 

For proposal generation, we first perform uniform downsampling with a large enough number of frames (\eg 256) and then treat each of the sampled frames as a proposal. 
While there are more sophisticated ways to generate proposals, including having multiple frames in a proposal, we find that the single frame proposal method works well with our top-K selection. Given a fixed number of frames $K$ to select, our single frame proposal ensures that each of the selected frames are selected because of its high relevance to the query. In contrast, if we generate proposals by segments (multiple consecutive frames), then there might be redundant information in the selected segments due to temporal proximity, which cannot make the best use of the budget $K$.

Given a query, we then use a LMM to score each proposal. Interestingly, we find that LMMs can give high quality scores without any training. We formulate scoring as a binary choice problem, which most LMMs are well-trained on. The exact prompt is included in \Cref{sec:supp_prompt} in supplementary, but the following prompt captures the main idea:

\begin{small}
\begin{verbatim}
Question: <query>
Does the image contain sufficient  
information to answer the given question?
A. yes, B. no
Answer with the option's letter.
\end{verbatim}
\end{small}

Most LMMs can follow the instruction and answer either ``A'' or ``B''. We use the probability that the LMM  answers ``A'' for a frame as its score. For a video frame, the score is often lower than 0.5 because it is less likely for a single frame to contain sufficient information to answer a question about the whole video. Nevertheless, the score is still indicative of the frame's relevance to the query.

\begin{table*}[t]
\centering
\tabsize
\begin{tabular}{ll|ccccc}
           &    & LongVideoBench                 & MLVU                           & VideoMME                       & EgoSchema                      & NeXT-QA                        \\
        LMM   &  Selection  & \textit{8min}       &  \textit{12min}  &  \textit{17min}                       & \textit{1.7min}              &  \textit{0.8min}                       \\\hline
LLaVA-OV-7B   & Uniform-32  & 54.2                           & 61.5                           & 56.7                           & 60.6                           & 78.1                           \\
              & FRAG-Top32-256 & 57.6                           & 65.3                           & 58.6                           & 61.1                           & 78.3                           \\
              &             & \cellcolor[HTML]{D9EAD3}+3.4\% & \cellcolor[HTML]{D9EAD3}+3.8\% & \cellcolor[HTML]{D9EAD3}+1.9\% & \cellcolor[HTML]{D9EAD3}+0.5\% & \cellcolor[HTML]{D9EAD3}+0.2\% \\\hline
LLaVA-OV-72B  & Uniform-32  & 61.0                           & 67.9                           & 65.8                           & 61.2                           & 82.7                           \\
              & FRAG-Top32-256 & 63.5                           & 69.5                           & 66.7                           & 62.7                           & 82.3                           \\
              &             & \cellcolor[HTML]{D9EAD3}+2.5\% & \cellcolor[HTML]{D9EAD3}+1.6\% & \cellcolor[HTML]{D9EAD3}+0.9\% & \cellcolor[HTML]{D9EAD3}+1.5\% & \cellcolor[HTML]{F4CCCC}-0.4\% \\\hline
InternVL2-8B  & Uniform-32  & 53.0                           & 59.8                           & 55.9                           & 54.6                           & 80.5                           \\
              & FRAG-Top32-256 & 57.0                           & 63.6                           & 57.1                           & 55.4                           & 80.6                           \\
              &             & \cellcolor[HTML]{D9EAD3}+4.0\% & \cellcolor[HTML]{D9EAD3}+3.9\% & \cellcolor[HTML]{D9EAD3}+1.2\% & \cellcolor[HTML]{D9EAD3}+0.8\% & \cellcolor[HTML]{D9EAD3}+0.1\% \\\hline
InternVL2-40B & Uniform-32  & 59.1                           & 63.6                           & 63.5                           & 61.5                           & 85.5                           \\
              & FRAG-Top32-256 & 62.6                           & 68.9                           & 65.7                           & 62.0                           & 84.6                           \\
              &             & \cellcolor[HTML]{D9EAD3}+3.5\% & \cellcolor[HTML]{D9EAD3}+5.2\% & \cellcolor[HTML]{D9EAD3}+2.2\% & \cellcolor[HTML]{D9EAD3}+0.4\% & \cellcolor[HTML]{F4CCCC}-0.9\% \\\hline
InternVL2-76B & Uniform-24  & 59.5                           & 63.3                           & 62.3                           & 63.1                           & 84.7                           \\
              & FRAG-Top24-256 & 61.5                           & 69.2                           & 66.0                           & 63.8                          &   84.5                             \\
              &             & \cellcolor[HTML]{D9EAD3}+2.0\% & \cellcolor[HTML]{D9EAD3}+5.8\% & \cellcolor[HTML]{D9EAD3}+3.7\% & \cellcolor[HTML]{D9EAD3}+0.7\% &  \cellcolor[HTML]{F4CCCC}-0.2\%     
\end{tabular}
\caption{Long video understanding results. \ours substantially improves over uniform sampling for long videos. Especially for LongVideoBench and MLVU which are curated to evaluate LMM's long video understanding. }
\label{tab:video}
\end{table*}

\subsection{Selection for Generation}

The next step is to select proposals based on their scores. We simply select the Top-K scoring proposals. 
The selected frames are then sorted by their temporal order for videos and page order for documents. Finally, the selected frames are fed into the answering LMM to generate the outputs. We follow the multi-image multimodal prompting format for each LMM to input the selected frames. 
We treat each frame in a video as an image, and do not use special formats for video frames. This allows us to use multi-image LLMs and not be limited to LMMs trained on videos.

We find that the simple Top-K selection works well for both videos and documents. For videos, although the scoring might not be perfect, selecting the Top-K gives room for errors. For example, in \Cref{fig:system}, one of the selected frames is focusing on a different woman. Nevertheless, the LMM can still ignore the irrelevant information and answer the question. Similarly for documents, although most of the information required to answer a question is contained in a single page, we can still select more than one page. Even if the scoring LMM does not give the highest score to the best page, it is more likely that it is in Top-K. 
While there are more complex approaches to select the frames based on the scores (\eg by considering the temporal diversity/proximity), our main message is that LMMs can zero-shot perform frame selection for long videos and long documents. Top-K selection is a direct way to demonstrate this.

\section{Experiments}
\label{sec:experiments}

\ours first selects relevant frames before generating the outputs. This is a simple yet effective approach for long videos and long documents. We thus evaluate both tasks.

\subsection{Evaluating Long Video Understanding}

\subsubsection{Datasets and Metrics} 

We consider five datasets: LongVideoBench (val, w/o sub)~\cite{wu2024longvideobench}, MLVU (dev)~\cite{MLVU}, Video-MME (w/o sub)~\cite{fu2024video}, EgoSchema (full)~\cite{mangalam2023egoschema}, and NeXT-QA (val)~\cite{xiao2021next}. LongVideoBench and MLVU are curated specifically to evaluate long video understanding in LMMs.  Video-MME contains short, medium, and long videos to provide a holistic view of video understanding. EgoSchema contains  long egocentric videos for evaluation. NeXT-QA is not curated for long video understanding but has been widely used in previous works. The average duration for the first three datasets are over 8 minutes, while Egoschema is around 2 minutes. NeXT-QA has the shortest average (0.8 minute). We report the question answering accuracy for all datasets.

\subsubsection{Implementation Details} 

\ours does not depend on a specific LMM. We consider two models, LLaVA-OneVision~\cite{li2024llava} (7B, 72B) and InternVL2~\cite{chen2024far} (8B, 40B, 76B), for evaluation. For all the videos, we first uniformly sample 256 frames. We then select the Top-32 frames except for InternVL2-76B, where we only select Top-24 due to memory limitation. For Video-MME, we use 28 frames only for InternVL2-40B for computation limitation when subtitles are included. 
In the main results we only report ``w/o sub'', but this allows direct comparison with ``w sub''.
During scoring, we use the default high resolution tiling for both models (Max 12 tiles for InternVL2 and AnyRes-9 for LLaVA-OneVIsion). During answering, high resolution tiling is disabled for both models for videos so that we can use the computational budget on as many distinct frames as possible. We follow the multi-image input template for both models to input the selected frames. No additional prompt is added to suggest that the images are selected from a video. We follow the official question prompting and evaluation for all datasets.

\begin{figure*}[t]
  \centering
  \includegraphics[width=1.0\linewidth]{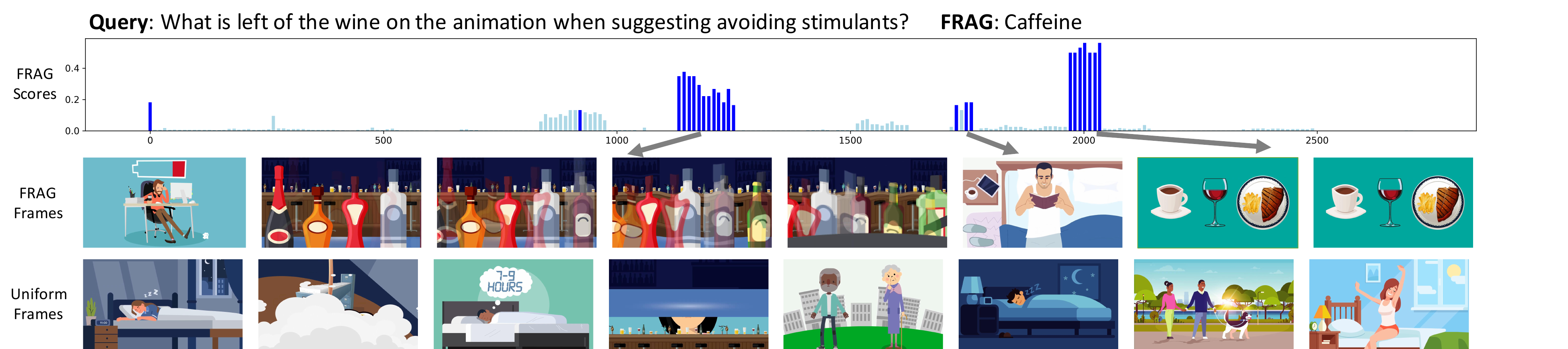}
  \caption{Qualitative result for \ours. x-axis is frame index, and y-axis is \ours score. \ours select frames that are much more relevant to the query and successfully answer the question. Uniform sampling misses the important frames and cannot answer the question.}
  \label{fig:qual_vid}
\end{figure*}

\begin{table*}[t]
\centering
\tabsize
\begin{tabular}{lcccccc}
               & Size & LongVideoBench & MLVU & VideoMME & EgoSchema & NeXT-QA \\ \hline
LongVA~\cite{zhang2024longva}         & 7B   & --             & 56.3 & 52.6     & --        & 68.3    \\
FRAG-LLaVA-OV  & 7B   & \textbf{57.6}           & \textbf{65.3} & \textbf{58.6}     & \textbf{61.1}      & 78.3    \\
FRAG-InternVL2 & 8B   & 57             & 63.6 & 57.1     & 55.4      & \textbf{80.6}    \\ \hline
Tarsier~\cite{wang2024tarsier}        & 34B  & --             & --   & --       & 61.7      & 79.2    \\
LLaVA-NeXT-Video~\cite{zhang2024llavanextvideo}& 34B  & 50.5          & --   & 52.0     & --        & 70.2    \\
VILA-1.5~\cite{lin2024vila}       & 40B  & --             & 56.7 & 62.3     & 58.7      & --      \\
FRAG-InternVL2 & 40B  & \textbf{62.6}           & \textbf{68.9} & \textbf{65.7}     & \textbf{62.0}      & \textbf{84.6}    \\ \hline
VideoLLaMA2~\cite{cheng2024videollama}    & 72B  & --             & 61.2 & 62.4     & \textbf{63.9}      & --      \\
FRAG-LLaVA-OV  & 72B  & \textbf{63.5}           & \textbf{69.5} & \textbf{66.7}     & 62.7      & 82.3    \\
FRAG-InternVL2 & 76B  & 61.5           & 69.2 & 66       & 63.8      &  \textbf{84.5}  \\ \hline
\multicolumn{7}{l}{\cellcolor[HTML]{EFEFEF}\emph{(Based on) Proprietary Models}}      \\
VideoAgent~\cite{wang2024videoagent}      &  --  & --             &  --  & --       & 54.1      & 71.3    \\
LVNet~\cite{park2024too}       &  --  &  --            &  --  & --       & 61.1      & 72.9    \\
VideoTree~\cite{wang2024videotree}      &  --  & --             &  --  & --       & 61.1      & 73.5    \\
GPT-4o         &  --  & 60.6           & 64.6 & 71.9     & --        & --     
\end{tabular}
\caption{\ours outperforms existing models of similar sizes. This includes long context model LongVA, and state-of-the-art 40B/76B models. \ours also outperforms methods that use GPT-4 and selection. Overall, \ours-LLaVA-OV performs better than \ours-InternVL2.}
\label{tab:video_sota}
\end{table*}

\subsubsection{Effectiveness of Frame Selection}
\label{sec:video}

The results for applying \ours on long video understanding are shown in \Cref{tab:video}. For each LMM, we compare Uniform-32 that uniformly samples 32 frames with \ours-Top32-256, which selects the top 32 frames out of 256 uniformly sampled frames using our approach. \ours substantially improves most of the results, especially on LongVideoBench and MLVU, which are designed specifically to evaluate long video understanding for LMMs. On Video-MME, which provides a more holistic view of video understanding, \ours still improves all of the evaluated LMMs. \ours also consistently improves on egocentric videos in EgoSchema. The only exception is NeXT-QA, \ours performs similarly to Uniform-32. Since the videos are less than a minute long in NeXT-QA, having 32 frames already captures most of the information in the video, and frame selection might not lead to improvements. 
Overall, we find that \ours can select better frames from videos, and consistently improve the performance for long videos. \Cref{fig:qual_vid} shows a qualitative result on a video with thousands of frames. The x-axis is the frame index, and y-axis is the score for each frame. Light blue shows scores for frames not selected, while blue shows scores for frames that are selected. \ours select frames that are much more relevant to the query and successfully answer the question, while uniform sampling misses the important frames.

\begin{table*}[t]
\centering
\tabsize
\begin{tabular}{ll|cccccc}
              &                          & \multicolumn{2}{c}{SlideVQA}                                      & \multicolumn{2}{c}{MP-DocVQA (val)}                                     & \multicolumn{2}{c}{MMLongBench-Doc}                               \\
LMM           & Selection                & EM                              & F1                              & Acc                             & ANLS                            & Acc                             & F1                              \\\hline
LLaVA-OV-7B   & Uniform-32 (1 tile)      & 43.0                            & 49.2                            & 28.7                            & 49.1                            & 13.6                            & 9.9                             \\
              & FRAG-Top2-All (16 tiles) & 59.8                            & 65.1                            & 66.6                            & 78.0                            & 23.7                            & 21.8                            \\
              &                          & \cellcolor[HTML]{D9EAD3}+16.8\% & \cellcolor[HTML]{D9EAD3}+15.9\% & \cellcolor[HTML]{D9EAD3}+37.9\% & \cellcolor[HTML]{D9EAD3}+28.9\% & \cellcolor[HTML]{D9EAD3}+10.0\% & \cellcolor[HTML]{D9EAD3}+11.9\% \\\hline
LLaVA-OV-72B  & Uniform-32 (1 tile)      & 48.9                            & 55.3                            & 31.6                            & 51.9                            & 19.7                            & 14.6                            \\
              & FRAG-Top2-All (16 tiles) & 69.6                            & 74.7                            & 74.5                            & 84.8                            & 32.8                            & 30.8                            \\
              &                          & \cellcolor[HTML]{D9EAD3}+20.7\% & \cellcolor[HTML]{D9EAD3}+19.4\% & \cellcolor[HTML]{D9EAD3}+42.9\% & \cellcolor[HTML]{D9EAD3}+32.9\% & \cellcolor[HTML]{D9EAD3}+13.2\% & \cellcolor[HTML]{D9EAD3}+16.2\% \\\hline
InternVL2-8B  & Uniform-32 (1 tile)      & 45.2                            & 53.4                            & 25.5                            & 40.6                            & 22.7                            & 14.8                            \\
              & FRAG-Top2-All (16 tiles) & 61.1                            & 67.7                            & 67.9                            & 77.9                            & 31.3                            & 27.0                            \\
              &                          & \cellcolor[HTML]{D9EAD3}+15.9\% & \cellcolor[HTML]{D9EAD3}+14.3\% & \cellcolor[HTML]{D9EAD3}+42.4\% & \cellcolor[HTML]{D9EAD3}+37.3\% & \cellcolor[HTML]{D9EAD3}+8.6\%  & \cellcolor[HTML]{D9EAD3}+12.3\% \\\hline
InternVL2-40B & Uniform-32 (1 tile)      & 54.8                            & 61.5                            & 37.5                            & 56.6                            & 26.2                            & 16.4                            \\
              & FRAG-Top2-All (16 tiles) & 71.9                            & 77.2                            & 79.5                            & 86.7                            & 38.4                            & 33.3                            \\
              &                          & \cellcolor[HTML]{D9EAD3}+17.1\% & \cellcolor[HTML]{D9EAD3}+15.7\% & \cellcolor[HTML]{D9EAD3}+42.0\% & \cellcolor[HTML]{D9EAD3}+30.1\% & \cellcolor[HTML]{D9EAD3}+12.2\% & \cellcolor[HTML]{D9EAD3}+16.9\% \\\hline
InternVL2-76B & Uniform-24 (1 tile)      & 57.1                            & 63.7                            & 40.1                            & 59.2                            & 24.7                            & 16.0                            \\
              & FRAG-Top2-All (12 tiles) & 72.7                            & 78.0                            & 80.5                            & 87.4                            & 37.9                            & 34.8                            \\
              &                          & \cellcolor[HTML]{D9EAD3}+15.6\% & \cellcolor[HTML]{D9EAD3}+14.3\% & \cellcolor[HTML]{D9EAD3}+40.4\% & \cellcolor[HTML]{D9EAD3}+28.3\% & \cellcolor[HTML]{D9EAD3}+13.2\% & \cellcolor[HTML]{D9EAD3}+18.8\%
\end{tabular}
\caption{Long document understanding results. \ours achieves even more significant improvements compared to long videos. For dense documents in MP-DocVQA, \ours achieves over 40\% improvements. Even for less dense slides in SlideVQA, \ours can still improve over 20\%. For the most challenging MMLongBench-Doc, \ours doubles the F1 scores of the uniform baseline.}
\label{tab:document}
\end{table*}

\subsubsection{Comparison with SOTA}

We have shown that \ours can select better frames and improve long video understanding. Now we compare \ours with state-of-the-art methods. The results are in \Cref{tab:video_sota}. For different model sizes, \ours models substantially outperform existing methods, and even outperform proprietary models like GPT-4o on LongVideoBench and MLVU. Overall, LLaVA-OV performs better than InternVL2. This is reasonable as LLaVA-OV has more training on video understanding. Nevertheless, \ours-InternVL2-40B still substantially outperforms all existing methods on all datasets. Existing long context models are limited to 7B size. \ours models of similar sizes already significantly outperforms LongVA, and we can easily extend to larger model sizes and still consider large numbers of sampled frames. Our largest 72/76B models also outperforms the recent SOTA model VideoLLaMA2-72B. Finally, we also compare to models that perform frame selection and grouping: VideoAgent, VideoTree, and LVNet. These methods first select or group video frames based on visual and textual information. The selected frames are then used to answer the question based on captioning models and proprietary LLMs, such as GPT-4. Without using proprietary models, \ours select high quality frames to answer the question for long videos. \ours-LLaVA-OV-7B already outperforms all these methods. Our larger models further improve the results.

\subsection{Evaluating Long Document Understanding}

\subsubsection{Datasets and Metrics} 

We consider three datasets: SlideVQA (test)~\cite{SlideVQA2023}, MP-DocVQA (val,test)~\cite{tito2023hierarchical}, and MMLongBench-Doc~\cite{ma2024mmlongbench}. SlideVQA contains slides of 20 pages and uses two metrics: exact match (EM) rate of the whole response and F1 score for words in the response. 
MP-DocVQA extends DocVQA~\cite{mathew2021docvqa} to multi-page. While its average page number (8.3) is smaller than SlideVQA (20), each page has more information. For metrics, MP-DocVQA uses accuracy (Acc) and Average Normalized Levenshtein Similarity (ANLS)~\cite{biten2019scene}. Acc measures exact matching, while ANLS provides a soft measure. Acc is only available for the validation set.
MMLongBench-Doc has the highest average number of pages (47.5). It has cross-page questions and unanswerable questions for evaluating hallucinations. For metrics, it uses accuracy and F1 score~\cite{ma2024mmlongbench}.

\subsubsection{Implementation Details} 

We use the same LMMs as long videos. For all documents, \ours goes through all the pages and selects the Top-2 pages. During scoring, we use the default high resolution tiling for both models (Max 12 tiles for InternVL2 and AnyRes-9 for LLaVA-OneVIsion). In contrast to videos, high resolution is crucial for document understanding. We thus use both AnyRes for LLaVA-OneVision and Dynamic Resolution for InternVL2 for documents during answering. We limit the number of tiles to 16 except InternVL2-76B, where we only use 12 tiles because of computational limitations. We follow the official question prompting and evaluation, and do not use additional prompts specific to \ours.

\subsubsection{Effectiveness of Frame Selection} 
\label{sec:document_effect}

The results for applying \ours on long document understanding are shown in \Cref{tab:document}. We again compare with the Uniform sampling baseline. We set the max number of tiles the same for \ours ($2 \times 16$) and Uniform ($32 \times 1$). 
For SlideVQA and MP-DocVQA, since the number of pages are often less than 32, this means that Uniform-32 in fact uses all the pages. As high resolution is crucial for document understanding, \ours models significantly outperform the uniform baselines. While text and information is less dense in slides, we still see up to 20\% improvements by focusing on the Top2 selected pages. For MP-DocVQA, where each page contains dense  information, we see up to 40\% improvements by using \ours. On the challenging MMLongBench-Doc, where both models have lower absolute performances, \ours still has strong relative improvements, and more than doubles the F1 scores of Uniform-32.

\subsubsection{Comparison with SOTA}

In this section, we compare \ours to existing methods. We use the test set for MP-DocVQA in this section. 

\vspace{1mm}
\noindent\textbf{SlideVQA.} The results are in \Cref{tab:slidevqa}. We compare with three existing approaches. M3D is a specialized model that uses the training set of SlideVQA to supervise not only the answer but also the page selection. BLIP-2 and InstructDr are LMMs finetuned on SlideVQA and visual document understanding datasets~\cite{InstructDoc2024}. Our 7B/8B models already improve F1 score by 16\% from the best existing methods (63.3 v.s. 47.3) without training on SlideVQA. Since \ours can be easily extended to larger LMMs, our best model (FRAG-InternVL2-76B) further improves by 12\% and leads to a total improvement of 28\% (75.6 v.s. 47.3).

\vspace{1mm}
\noindent\textbf{MP-DocVQA.} The results are in \Cref{tab:mpdocvqa}. We compare with multi-image LMMs (LLaVA-NeXT-Interleave~\cite{li2024llavainter}), long context LMMs (LonVA~\cite{zhang2024longva}, Idefics3~\cite{laurençon2024building}), and LMMs specialized for multi-page documents (mPLUG-DocOwl-2~\cite{hu2024mplug}). While mPLUG-DocOwl-2 outperforms other existing LMMs, our 7B/8B models further improve the ANLS by 10\% (79.1 vs 69.4). Note that mPLUG-DocOwl-2 trains on MP-DocVQA, while \ours is zero-shot. Our best model (FRAG-InternVL2-76B) further improves the ANLS by 9\% and is currently the \#1 method on the MP-DocVQA leaderboard, which includes several supervised specialist models for multi-page documents~\cite{blau2024gram,tito2023hierarchical}. 

\begin{table}[t]
\centering
\begin{tabular}{lcccc}
               & Size & Zero-shot & EM   & F1   \\\hline
M3D~\cite{SlideVQA2023}            & --   & \xmark    & 33.5 & 41.7 \\
BLIP-2~\cite{li2023blip}         & 3.4B & \xmark    & 36.9 & 46.5 \\
InstructDr~\cite{InstructDoc2024}     & 3.4B & \xmark    & 37.7 & 47.3 \\
FRAG-LLaVA-OV  & 7B   & \cmark    & 51.5 & 60.7 \\
FRAG-InternVL2 & 8B   & \cmark    & \textbf{52.9} & \textbf{63.3} \\\hline
FRAG-InternVL2 & 40B  & \cmark    & 63.5 & 73.0 \\
FRAG-LLaVA-OV  & 72B  & \cmark    & 64.5 & 73.3 \\
FRAG-InternVL2 & 76B  & \cmark    & \textbf{66.0} & \textbf{75.6}
\end{tabular}
\caption{SlideVQA results. Our smallest models already zero-shot outperform supervised specialist models. Our largest models further improve and lead to a 28\% improvement.}
\label{tab:slidevqa}
\end{table}

\begin{table}[t]
\centering
\begin{tabular}{lcc}
               & Size & ANLS \\\hline
LLaVA-NeXT-Interleave~\cite{li2024llavainter}         & 7B       & 44.9 \\
LongVA~\cite{zhang2024longva}         & 7B       & 60.8 \\
Idefics3~\cite{laurençon2024building}         & 8B       & 67.2 \\
mPLUG-DocOwl-2~\cite{hu2024mplug} & 8B       & 69.4 \\
FRAG-LLaVA-OV  & 7B       & \textbf{79.1} \\
FRAG-InternVL2 & 8B       & 77.8 \\\hline
FRAG-InternVL2 & 40B      & 86.6 \\
FRAG-LLaVA-OV  & 72B      & 85.0 \\
FRAG-InternVL2 & 76B      & \textbf{88.3}
\end{tabular}
\caption{\ours MP-DocVQA results. \ours outperforms existing multi-page document LMMs of similar sizes. Our 76B model further improves and leads to almost 20\% improvements. }
\label{tab:mpdocvqa}
\end{table}

\begin{table}[t]
\centering
\begin{tabular}{lccc}
                 & Size & Acc  & F1   \\\hline
InternLM-XC2-4KHD~\cite{dong2024internlm} & 8B & 10.3 & 9.8 \\
mPLUG-DocOwl-1.5~\cite{hu2024mplug15} & 8B   & 6.9  & 6.3  \\
FRAG-LLaVA-OV    & 7B   & 23.7 & 21.8 \\
FRAG-InternVL2   & 8B   & \textbf{31.3} & \textbf{27.0} \\\hline
InternVL-1.5~\cite{chen2024far}     & 26B  & 14.6 & 13.0 \\
FRAG-InternVL2   & 40B  & \textbf{38.4} & 33.3 \\
FRAG-LLaVA-OV    & 72B  & 32.8 & 30.8 \\
FRAG-InternVL2   & 76B  & 37.9 & \textbf{34.8} \\\hline
GPT-4V           & --   & 32.5 & 31.4 \\
GPT-4o           & --   & 40.8 & 42.7
\end{tabular}
\caption{MMLongBench-Doc results. \ours substantially outperforms existing opensource LMMs, and our best model \ours-InternVL2-76B even outperforms GPT-4V.}
\label{tab:mmlbdoc}
\end{table}

\vspace{1mm}
\noindent\textbf{MMLongBench-Doc.} The results are in \Cref{tab:mmlbdoc}. We first compare with InternLM-XC2-4KHD~\cite{dong2024internlm}, which is the best performing LMM below 14B. Our FRAG-InternVL2-8B almost tripled both its Acc and F1 score. We also include mPLUG-DocOwl-1.5 for reference as we also compare with mPLUG-DocOwl-2 on MP-DocVQA. Currently, the best performing opensource LMM is InternVL-1.5. \ours's results more than double both of its metrics. Our best models (FRAG-InternVL2-40B/76B) even outperforms GPT-4V. This is a challenging benchmark that contains documents that average almost 50 pages. For our baselines in \Cref{sec:document_effect}, we use InternVL-2, which does outperform InternVL-1.5. Nevertheless, \ours still significantly outperforms this stronger baseline.

Overall, \ours significantly outperforms all existing methods on all three benchmarks. This shows the effectiveness of our frame selection framework for long document understanding. We find that InternVL2 generally works better than LLaVA-OV for document understanding. This is in contrast to videos, where LLaVA-OV performs better.

\begin{table*}[t]
\centering
\tabsize
\begin{tabular}{ll|ccccc|c}
Answering LMM & Scoring LMM    & LongVideoBench & MLVU & VideoMME & EgoSchema & NeXT-QA & Avg \\\hline
InternVL2-40B (32) & None (Uniform) & 59.1           & 63.6 & 63.5     & 61.5      & \textbf{85.5} & 66.7    \\
InternVL2-40B (32) & SigLIP~\cite{zhai2023sigmoid}         & 61.5           & 67.7 & 62.6     & 58.2      & 82.0  & 66.8  \\
InternVL2-40B (32) & InternVL2-1B   & 60.4           & 67.9 & 63.6     & 61.2      & 84.3  & 67.5   \\
InternVL2-40B (32) & InternVL2-8B   & 62.5           & {\ul69.9} & 64.3     & 61.7      & {\ul85.1} & 68.7    \\
InternVL2-40B (32) & InternVL2-40B  & {\ul62.6}           & 68.9 & \textbf{65.7}     & {\ul62.0}      & 84.6 & {\ul68.8}   \\
InternVL2-40B (32) & InternVL2-76B  & \textbf{63.4}           & \textbf{70.3} & {\ul65.4}     & \textbf{62.3}      &  84.6  & \textbf{69.2}   
\end{tabular}
\caption{Results with different scoring LMMs. We find that larger models thus better multimodal understanding lead to better frame selection. SigLIP hurts on three out of five benchmarks, and the average is similar to uniform sampling.}
\label{tab:scoring}
\end{table*}

\begin{figure*}[t]
  \centering
  \includegraphics[width=1.0\linewidth]{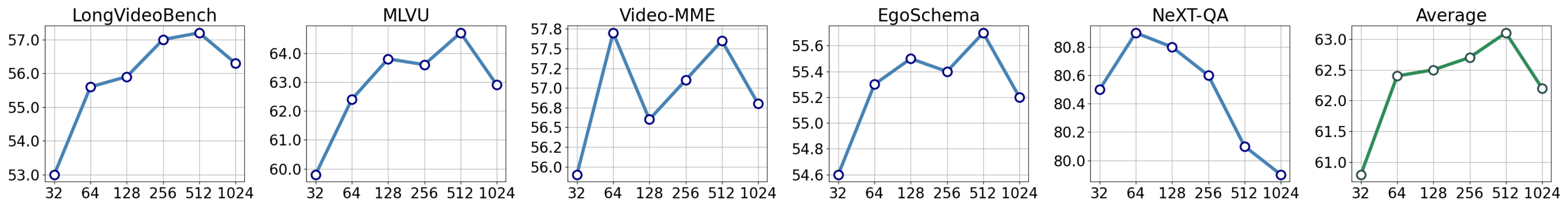}
  \caption{Results with different numbers of sampled frames. Performance peaks at 512 frames, and regress at 1024 frames for our setting. Oversampling can lead to concentrated Top-K frames with less diversity and hurt performance.}
  \label{fig:sample_ablation}
\end{figure*}

\begin{figure*}[h!]
  \centering
  \includegraphics[width=1.0\linewidth]{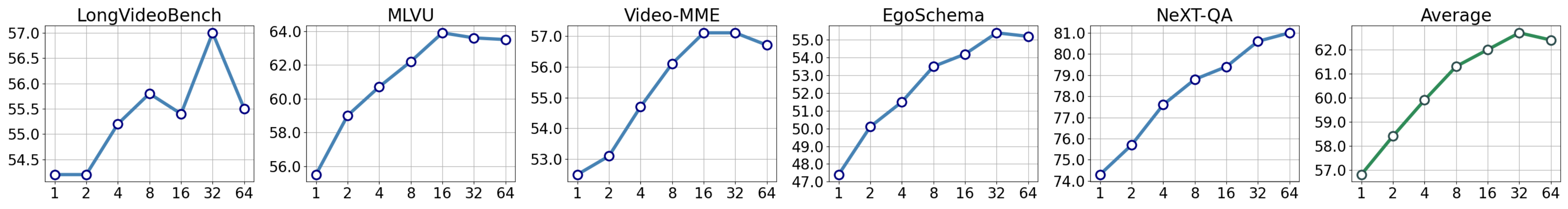}
  \caption{Results with different numbers of selected frames. This is also the number of input frames to the answering LMM. Performance peaks at 32 frames, which is consistent with findings in other works. We thus use 32 frames for our experiments}
  \label{fig:input_ablation}
\end{figure*}

\subsection{Ablation Studies}

\subsubsection{Evaluating Scoring LMM.}

First we analyze the effect of different scoring LMMs. In previous experiments, we use the same LMM for both scoring and answering. However, the scoring LMM and the answering LMM do not have to be the same. In this section, we analyze the effect of having different scoring LMMs given the same answering LMM. The results are in \Cref{tab:scoring}. We use InternVL2-40B as the answering LMM and vary the size of scoring LMM. We additionally include SigLIP~\cite{zhai2023sigmoid} to evaluate the importance of using large language models. We also show the uniform sampling baseline as reference. 

On average, using SigLIP is marginally better than uniform sampling. However, it hurts the performances on three out of the five benchmarks. The smallest model variant, InternVL2-1B slightly improves over SigLIP, and does not hurt performance as much on shorter videos in EgoSchema and NeXT-QA (over 3\% drop for SigLIP). For the rest of the models, we see a positive trend as we increase the model size. InterVL2-76B is the best scoring LMM even when we use InternVL2-40B as the answering LMM. This shows that the frame selection quality correlates with the general multimodal capabilities of LMM, and the selected frames can transfer between different models.

\subsubsection{Evaluating Number of Sampled Frames} 
\label{sec:sample}

Recall that for a long video, we do not score every frame in the video. Instead, we first perform uniform downsampling, and only score the sampled frames. The results for sampling different numbers of frames are in \Cref{fig:sample_ablation}. The x-axis is the number of sampled frames. We fix $K=32$ for Top-K selection. We use InternVL2-8B for answering and scoring.

From the average plot, the performance improves as we increase the number of sampled frames until 512 frames, and the performance regresses for 1024 frames. 
We compare the frames selected by FRAG-Top32-512 and FRAG-Top32-1024 and find that since we use a simple Top-K selection without any constraint, the Top-K frames could be too concentrated in a short time period if too many frames are sampled. 
This is a limitation of our simple approach and could be improved by more advanced selection approaches (\eg by considering frame diversity). Nevertheless, we believe our current approach could still illustrate the benefit of using strong LMMs for frame selection. We provide further analysis of this phenomenon in \Cref{sec:supp_sample} in supplementary. 

Most benchmarks follow the same trend (\ie peaking at 512) except NeXT-QA, which peaks at 64 sampled frames. This is due to a similar reason, where oversampling frames leads to concentrated selection. NeXT-QA has the shortest average video length (0.8min), and thus even 128 frames is oversampling. This also explains the performance of \ours on NeXT-QA in \Cref{sec:video}. For the main experiments, we use 256 sampled frames for all datasets.

\subsubsection{Evaluating Number of Selected Frames}

In \Cref{sec:sample}, we change the number of sampled frames. Now we fix the number of sampled frames to 256 as in our main experiments, and evaluate the change of $K$ in Top-K selection. This is also the number of input frames for answering. We again use InterVL2-8B. The results are in \Cref{fig:input_ablation}. Based on the average, the overall trend is that the performance continues to improve until 32 frames. Similar effect is also observed by Wu \etal~\cite{wu2024longvideobench} on several other open source models, where the model performance collapses after too many frames are used as input. We thus use 32 frames in our main experiments.

\section{Conclusion}
\label{sec:conclusion}

We propose Frame Selection Augmented Generation (\ours), which processes long inputs without long context LMMs. \ours substantially improves long video and long document understanding by first selecting the frames or pages that are relevant to the query. We demonstrate that this can be achieved zero-shot without fine-tuning, and only using a simple Top-K selection. Our experiments show that this unified and minimal framework achieves state-of-the-art results on both long video and long document understanding. We further analyze the effect of using different LMMs for selection and show that stronger LMMs do lead to better selection, and frames selected by weak models in fact degrade the performance. Future works include better selection strategy and improving the efficiency of \ours.

{
    \small
    \bibliographystyle{ieeenat_fullname}
    \bibliography{main}
}

\clearpage
\setcounter{page}{1}
\maketitlesupplementary

\setcounter{section}{0}
\renewcommand{\thesection}{\Alph{section}}

\begin{figure*}[t]
  \centering
  \includegraphics[width=1.0\linewidth]{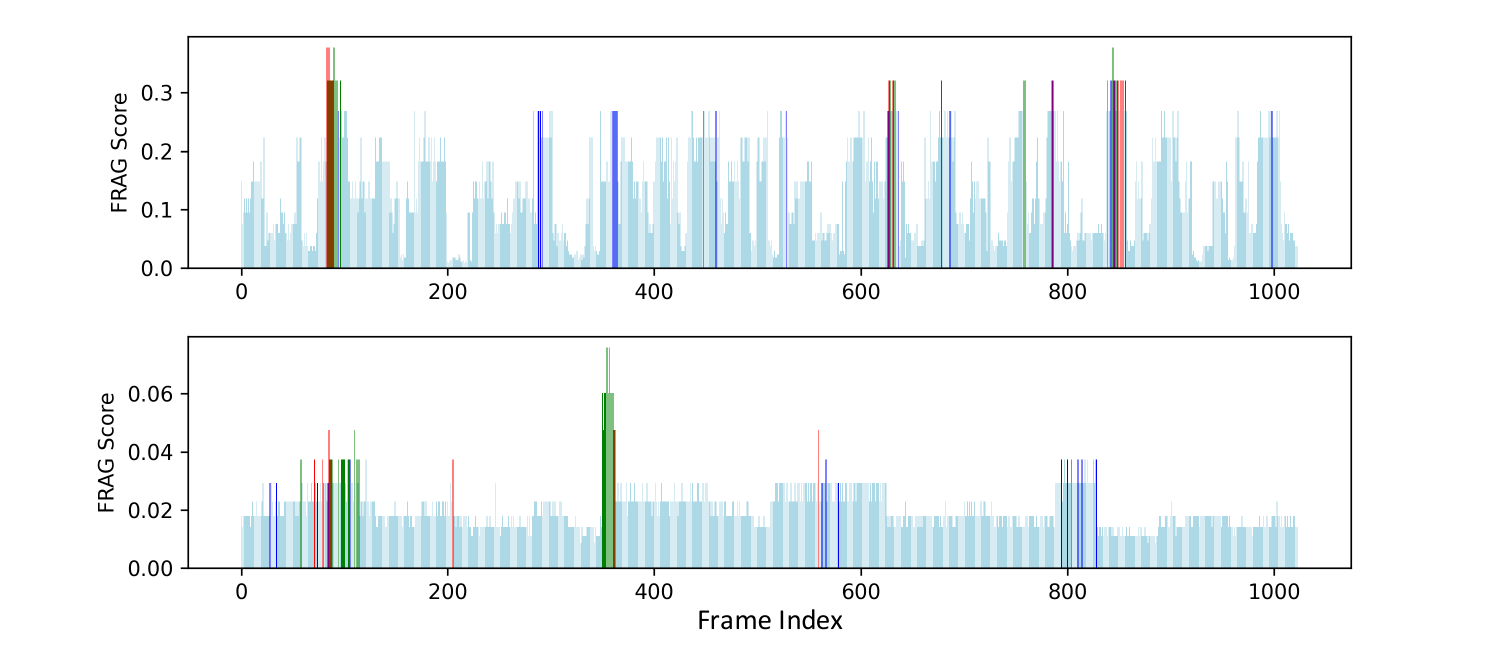}
  \caption{Visualization of videos where FRAG-Top32-512 gives the correct answer, but FRAG-Top32-1024 gives the wrong answer. Blue bars are frames selected by FRAG-Top32-512 but not FRAG-Top32-1024. The temporally diverse blue bars show that FRAG-Top32-512 selects more diverse frames, which lead to the correct answers.}
  \label{fig:supp1}
\end{figure*}

\begin{figure*}[t]
  \centering
  \includegraphics[width=1.0\linewidth]{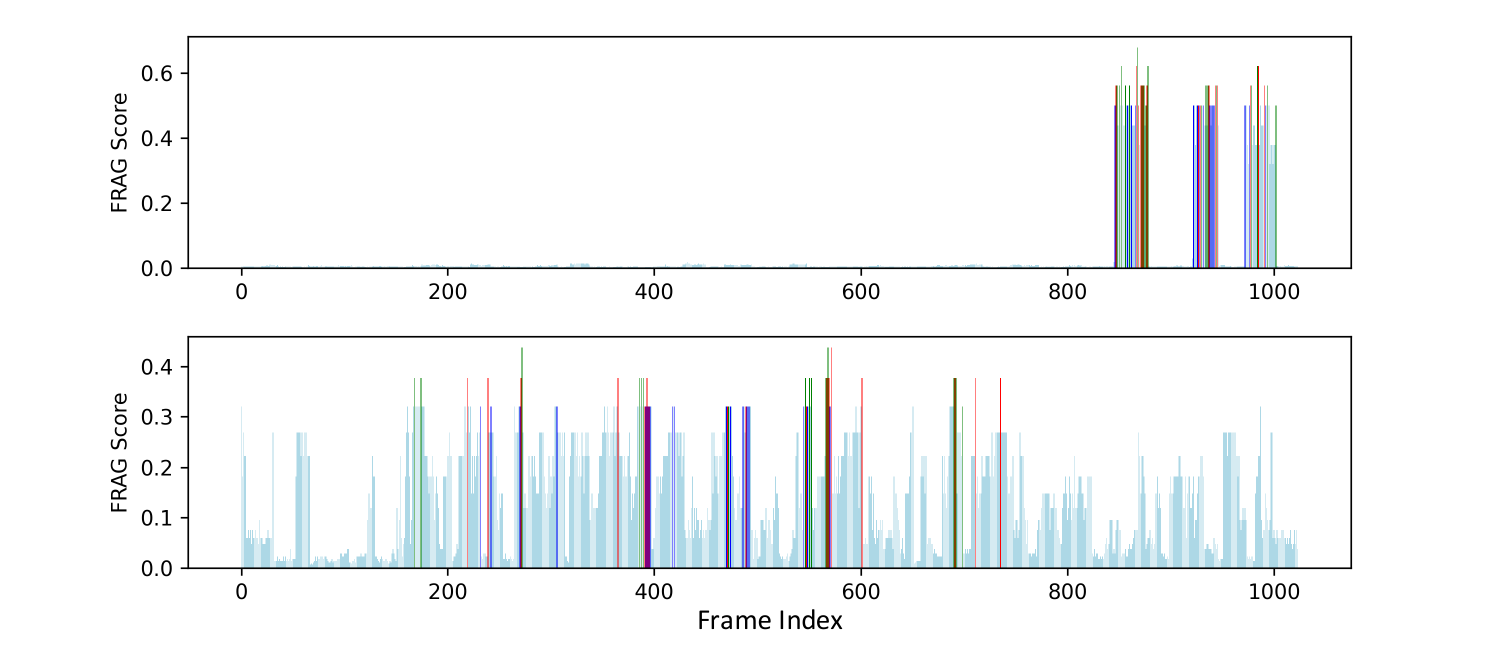}
  \caption{Visualization of videos where FRAG-Top32-1024 gives the correct answer, but FRAG-Top32-512 gives the wrong answer. For these videos, the diversity advantage is less obvious.}
  \label{fig:supp2}
\end{figure*}

\section{Selection Scoring Prompt}
\label{sec:supp_prompt}

Our prompt is based on the prompt used in SeViLA~\cite{yu2024self}. The full prompt for scoring is the following:
\begin{small}
\begin{verbatim}
Question: <query>
Does the information within the image
provide the necessary details to 
accurately answer the given question?
A. yes
B. no
Answer with the option's letter from
the given choices directly.
\end{verbatim}
\end{small}

\section{Analysis of Number of Sampled Frames}
\label{sec:supp_sample}

In \Cref{sec:sample}, we evaluate the effect of different numbers of sampled frames (\ie FRAG-Top32-$N$). Interestingly, we find that while initially the performance improves as we sample more frames, it starts to regress as we sample too many frames. The explanation is that since we use a simple Top-K selection without any constraint, the Top-K frames could be too concentrated in a short time period if too many frames are sampled. Now we visualize this phenomenon by comparing FRAG-Top32-512 and FRAG-Top32-1024 on LongVideoBench. First, in \Cref{fig:supp1}, we visualize videos where FRAG-Top32-512 gives the correct answer, while FRAG-Top32-1024 gives the wrong answer. These are the videos where the regression happens. 
The x-axis is the frame index, and the y-axis is the FRAG score for each frame. We only show the scores of the 1024 sampled frames in FRAG-Top32-1024, which is a superset of frames sampled by FRAG-Top32-512. 
Light blue bars are frames that are selected by neither FRAG-Top32-512 nor FRAG-Top32-1024. Green bars are frames that are selected by both FRAG-Top32-512 and FRAG-Top32-1024. Blue bars are frames that are selected by FRAG-Top32-512 but not FRAG-Top32-1024. Red bars are frames that are selected by FRAG-Top32-1024 but not FRAG-Top32-512. The main things to focus on are the blue bars. These frames are selected by FRAG-Top32-512 but not FRAG-Top32-1024, which most likely cause the change of model outputs. We can see that blue bars in these videos spread more diversely across the whole video, while red and green bars are more concentrated at nearby frames. Blue bars are more temporally diverse because there are less sampled frames in highest scoring segments. Thus, not all Top-32 frames are concentrated in the highest scoring segments, which in turn lead to more diverse selection. On the other hand, in \Cref{fig:supp2}, we visualize  videos where FRAG-Top32-1024 gives the correct answer, while FRAG-Top32-512 gives the wrong answer. Here, the diversity advantage is less obvious.

\end{document}